\documentclass{article}
\usepackage{spconf,amsmath,graphicx,amssymb}
\usepackage{booktabs}
\usepackage{tabularx}
\usepackage{color}
\usepackage[colorlinks,bookmarks=true,citecolor=black,linkcolor=black,urlcolor=blue]{hyperref}

\let\OLDthebibliography\thebibliography
\renewcommand\thebibliography[1]{
  \OLDthebibliography{#1}
  \setlength{\parskip}{0pt}
  \setlength{\itemsep}{0pt plus 0.3ex}
}

\begin{document}\sloppy

% Example definitions.
% --------------------
\def\x{{\mathbf x}}
\def\L{{\cal L}}

% Title.
% ------
\title{Skipformer: A Skip-and-Recover Strategy for Efficient Speech Recognition}
%
% Single address.
% ---------------
% \name{Anonymous ICME submission}
\address{Building 4 West Block Yard 10 Xibeiwang East Rd Haidian Dist Beijing, 100089 China}
%Address and e-mail should NOT be added in the submission paper. They should be present only in the camera ready paper. 
\name{Wenjing Zhu, Sining Sun, Changhao Shan, Peng Fan, Qing Yang\sthanks{Corresponding author.}}
\address{Du Xiaoman AI Lab, Beijing, China \\ \{zhuwenjing02, sunsining, shanchanghao, fanpeng, yangqing\}@duxiaoman.com}

%
% For example:
% ------------
%\address{School\\
%	Department\\
%	Address}
%
% Two addresses (uncomment and modify for two-address case).
% ----------------------------------------------------------
% \twoauthors
%  {A. Author-one, B. Author-two\sthanks{Thanks to XYZ agency for funding.}}
% 	{School A-B\\
% 	Department A-B\\
% 	Address A-B}
%   {C. Author-three, D. Author-four\sthanks{The fourth author performed the work
% 	while at ...}}
% 	{School C-D\\
% 	Department C-D\\
% 	Address C-D}
%
% \begin{document}
%\ninept
%
\maketitle
\begin{abstract}

Conformer-based attention models have become the \textit{de facto} backbone model for Automatic Speech Recognition tasks. 
A blank symbol is usually introduced to align the input and output sequences for CTC or RNN-T models. 
Unfortunately, the long input length overloads computational budget and memory consumption quadratically by attention mechanism. 
In this work, we propose a ``Skip-and-Recover" Conformer architecture, named Skipformer, to squeeze sequence input length dynamically and inhomogeneously. 
Skipformer uses an intermediate CTC output as criteria to split frames into three groups: crucial, skipping and ignoring. 
The crucial group feeds into next conformer blocks and its output joint with skipping group by original temporal order as the final encoder output. 
Experiments show that our model reduces the input
sequence length by 31 times on Aishell-1 and 22 times on Librispeech corpus.
Meanwhile, the model can achieve better recognition accuracy and faster inference speed than recent baseline models. 
Our code is available at https://github.com/Duxiaoman-DI/public-achievements-on-Speech/tree/skipformer.
\end{abstract}
\begin{keywords}
speech recognition, complexity reduction, end-to-end
\end{keywords}
\section{Introduction}\label{sec:intro}
In recent years, end-to-end~(E2E) models~\cite{graves2006connectionist,AED,LAS,graves2012sequence} have brought significant progress in the field of automatic speech recognition~(ASR). 
There are three popular E2E models for ASR tasks: Connectionist Temporal Classification~(CTC)~\cite{graves2006connectionist}, Attention-based Encoder-Decoder (AED)~\cite{AED,LAS} and Recurrent Neural Network Transducer~(RNN-T)~\cite{graves2012sequence}. 
In E2E ASR models, the most important component is the acoustic encoder which converts speech input sequences into high-level feature representations. 
Thanks to the development of the attention mechanism, Conformer-based encoder has been proven to outperform other network architectures and got state-of-the-art performance on many ASR tasks. 
However, Conformer-based encoder suffers from the quadratic complexity of the attention mechanism limiting its efficiency on long sequence lengths. 
Especially for the non-streaming AED model, this issue is more serious because an extra cross-attention needs to be calculated between encoder and decoder.  
On the other hand, the length of input sequence is much longer than that of the output sequence.
In order to align the input and output sequences, CTC and RNN-T models introduce an extra blank symbol. 
For CTC or RNN-T models, blank symbol dominates the predicted output sequences, which makes a little contribution to the final performance and results in redundant computation.  

Many efforts have been devoted to improving the computation efficiency of acoustic encoders. 
The most recent approaches is using bigger downsampling factors to decrease sequence length.  
The vanilla Conformer~\cite{conformer} encoder begins  with two convolution layers with stride 2 respectively along temporal dimension to reduce the input sequence length by 4 times. 
Efficient Conformer~\cite{Efficientconformer} increased the downsamping factor to 8 by using progressive downsampling method without performance drop. 
Squeezeformer~\cite{squeezeformer} combines downsampling with temporal U-Net structure. Followed by original down-sampling schema, an extra downsampling layer is applied at the middle of encoder layer and upsampling layer is applied at the end of encoder layer to recover 4 times temporal resolution to keep the recognition accuracy. 
A similar strategy was also used for deeper downsampling in Uconv-Conformer~\cite{Uconvconformer}. 

As for RNN-T and CTC model, blank symbols can be ignored during decoding, which can accelerate the decoding process \cite{blankregularized,wang2023accelerating,FSR,factorized}. Furthermore, the CTC output can also be used to guide the encoder downsampling. For example, in~\cite{wang2023accelerating}, they proposed to do frame reduction in the middle of RNN-T encoder using co-trained CTC guidance. Compared with downsampling frames uniformly, such as the Efficient Conformer and Squeezeformer, sampling with CTC guidance is a kind of ``importance sampling'' method. 
Blank Frames are dropped directly by such a CTC guidance. 
Similarly, we try to introduce the CTC guidance method into AED models to drop blank frames.
Unfortunately, the performance drops greatly after decoding without blank symbols. 

Building upon this opinion, in this paper, we propose a ``Skip-and-Recover'' Conformer architecture, named Skipformer. Our core idea is that the less useful information one frame contains, the simpler model required to model it. 
On the contrary, the more crucial information one frame contains, the more complex model required to model it. 
In this work, an intermediate CTC output is used as criteria to measure the importance of an acoustic frame. Specifically, an intermediate CTC loss is attached to layer $M$th of the Conformer blocks which totally have $(M+N)$ layers. 
Frames after layer $M$ will be split into three groups guided by an intermediate CTC: the crucial group, trivial group and ignoring group.  
In general, the crucial group contains the most of non-blank semantic frames. 
Correspondingly, the trivial group and ignoring group contain the most of blank frames. 
The tivial frames serve as boudaries to patition repeat symbols. 
And the rest frames are classified into ignoring group, which will be discarded directly. 
These frames from the crucial group will be fed into the next $N$ Conformer layers to extract more valuable information by using more layers. 
As for the frames from the trivial group, they will skip over the following $N$ layers. 
At the end of the Conformer encoder, frames from crucial and skipping groups will be jointed together by original sequential order. 
Finally, the encoder output is fed into attention decoder for decoding. 
Furthermore, we explored several different grouping strategies to find out how to classify blank frames into trival and ignoring groups. 
Experimental results show that our proposed Skipformer can obtain better recognition performance and reduce the inference time significantly. 

\section{Skipformer}\label{sect_model}
\begin{figure}[tb!]
  \centering
  \includegraphics[width=\columnwidth]{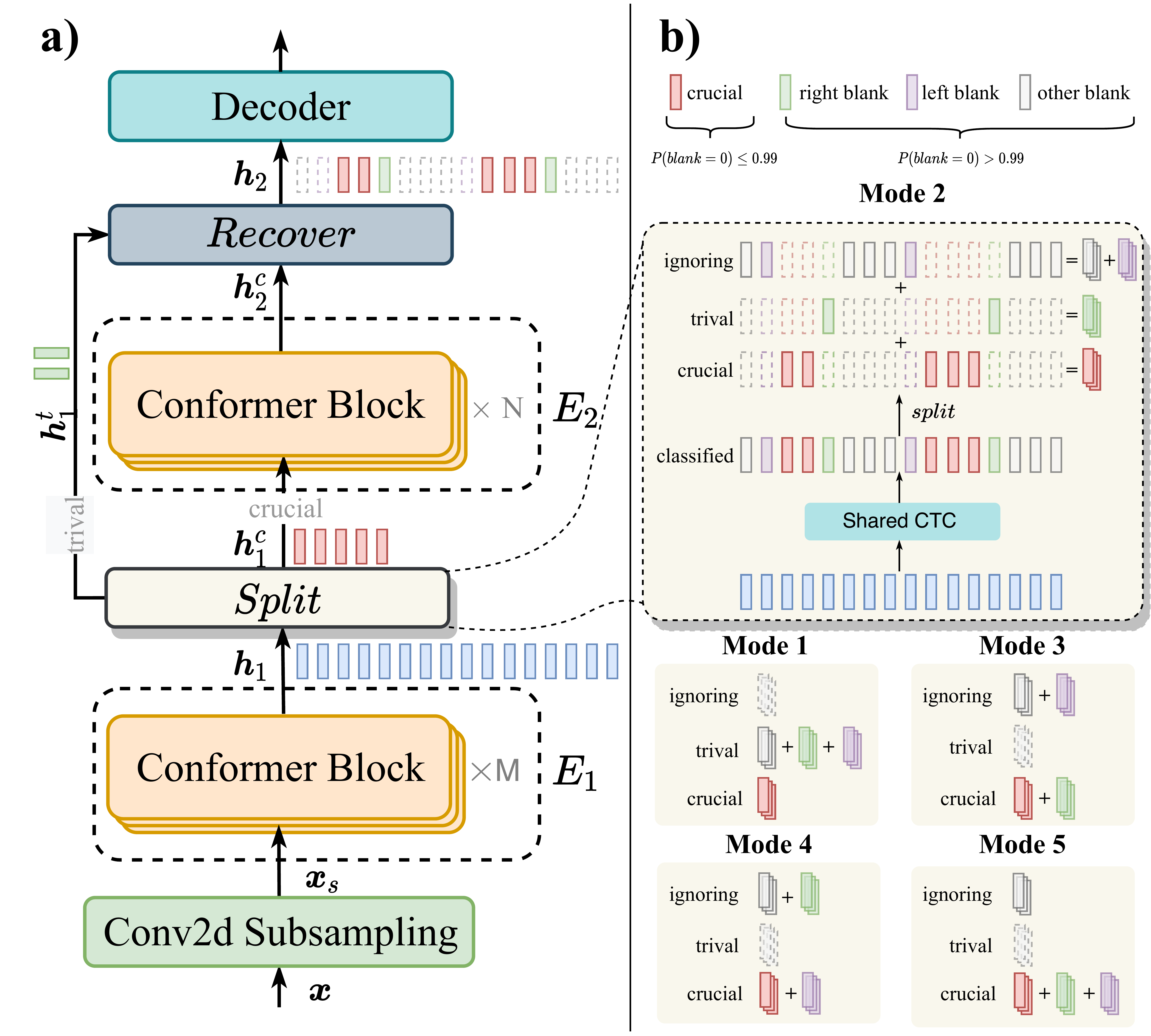}
  \caption{Overview of the model architecture. a) model architecture; b) split mode. \label{fig:model}}
\end{figure}

In this section, we present our Skipformer model in detail. 
Figure \ref{fig:model} a) shows the Encoder-Decoder architecture of Skipformer with split Mode 2. 
Our core idea is that the less useful information one frame contains, the simpler model required to model it. 
On the contrary, the more crucial information one frame contains, the more complex model required to model it. 
Our proposed Skipformer consists of two main operations: frame skipping and recovering. 
The ``Skip-and-Recover" strategy is a trade-off between computation complexity and recognition accuracy. 
Thanks to the nice property of the CTC loss function, we classify frames to be blank and none-blank by using an intermediate CTC output probabilities. 
With different strategies, the labeled frames are classified to three groups as mentioned above. 

\subsection{Overview of Skipformer}
As shown in Fig.~\ref{fig:model} a), our Skipformer adopts the joint CTC and AED structure such as U2~\cite{u2}. As for encoder, the vanilla Conformer encoder architecture with a Conv2d subsampling frontend is used.  
We split the Conformer encoder into two sub-encoders $E_1$ and $E_2$ with $M$ and $N$ Conformer blocks. 
The output of encoder \(E_1\) and \(E_2\) are fed into shared decoder to compute losses. 
Given input sequence $\boldsymbol{x}=\{x_0, x_1, ..., x_{T-1}\}$, the formula of our proposed Skipformer can be described as equations (1) to (5).

Among them, equations (3) to (5) depict the Skip-and-Recover strategy. 
First, in equation (3), the output sequence $\boldsymbol{h}_1$ of encoder $E_1$ is split into three sub-sequences according to the intermediate CTC output.  
$\boldsymbol{h}_1^c$, which belongs to crucial group, consists of semantic features for ASR task.   
And $\boldsymbol{h}_1^t$ contains trivial frames that need to be skipped over $E_2$ to output directly. 
While frames in $\boldsymbol{h}_1^i$ belongs to ignoring group and is ignored directly in the following calculation. 
Secondly, in equation (4) only $\boldsymbol{h}_1^c$ will be fed into encoder $E_2$ and get the output  $\boldsymbol{h}_2^c$. 
Because $\boldsymbol{h}_1^c$ is shorter than  $\boldsymbol{h}_1$, our proposed model is much computationally efficient. 
Finally, the $Recover$ function combines  $\boldsymbol{h}_2^c$ and  $\boldsymbol{h}_1^t$ together according to their original time order. 
The result sequence  $\boldsymbol{h}_2$ has shorter sequence length by dropping ignoring frames and keeps the temporal information in order. 

\begin{align}
\boldsymbol{x}_s & = Conv2dSubsampling(\boldsymbol{x}) \\
\boldsymbol{h}_1 & = E_1(\boldsymbol{x}_s) \\
\boldsymbol{h}_1^c, \boldsymbol{h}_1^t, \boldsymbol{h}_1^i  & = Split(\boldsymbol{h}_1|CTC(\boldsymbol{h}_1)) \\
\boldsymbol{h}_2^c & = E_2(\boldsymbol{h}_1^c) \\
\boldsymbol{h}_2 & = Recover(\boldsymbol{h}_2^c,  \boldsymbol{h}_1^t)
\end{align}
\begin{table}[bt!]
  \caption{ Frames contain in $(\boldsymbol{h}_1^c, \boldsymbol{h}_1^t, \boldsymbol{h}_1^i)$ of each Mode. $\mathbb{C}$ is used to represent set of frames that have non-blank intermediate CTC output.  Set $\mathbb{L}$ or $\mathbb{R}$ contains the nearest left or right blank frame of each non-blank frame in $\mathbb{C}$ . Set $\mathbb{B}$ contains all blank frames.  }\label{tab:mode}
  \centering
  \begin{tabular}{cccc}
    \toprule
    Mode & \(\boldsymbol{h}_1^c\) & \(\boldsymbol{h}_1^t\)  & \(\boldsymbol{h}_1^i\)  \\
    \midrule
    1  & $\mathbb{C}$ & $\mathbb{B}$  & $\emptyset$ \\
    2 & $\mathbb{C}$ & $\mathbb{R}$  & $\mathbb{B}$ $\setminus \mathbb{R}$ \\
    3 & $\mathbb{C}\cup \mathbb{R}$ & $\emptyset$ & $\mathbb{B}\setminus \mathbb{R}$ \\
    4 & $\mathbb{L}\cup \mathbb{C}$ & $\emptyset$ & $\mathbb{B}\setminus \mathbb{L}$ \\
    5 & $\mathbb{L} \cup \mathbb{C}\cup \mathbb{R}$ & $\emptyset$ &  $\mathbb{B}\setminus \mathbb{R} \setminus \mathbb{L}$\\
    
    \bottomrule
  \end{tabular}
\end{table}

\subsection{Data Split Strategy}
Although non-blank frames are cruical to represent semantic features, 
We use encoder $E_2$ further to extract high level features. 
Experiments show worse performance by removing all blank frames before decoding. 
The reason might be that blank symbols play an important role to seperate semantic tokens and merge repeat symbols. 
Since the transformer decoder also suffers from quadratical computaition complexity. 
Shorter sequence length can accelerate the computation of both encoder and transformer decoder. 
Hence, as shown in Fig.~\ref{fig:model} b), we propose five kinds of strategies to investigate which frames can be ignored or skipped without performance drop. 
Table~\ref{tab:mode} describes the set of frames from groups of $(\boldsymbol{h}_1^c, \boldsymbol{h}_1^t, \boldsymbol{h}_1^i)$ in each mode. 
We use $\mathbb{C}$ to represent set of frames that have non-blank symbols. 
Set $\mathbb{L}$ or $\mathbb{R}$ contains the nearest left or right blank frame of each non-blank frame in $\mathbb{C}$ . 
Set $\mathbb{B}$ contains all blank frames. 
$\boldsymbol{h}_1^c$ of   Mode 1 and Mode 2 only contains non-blank frames, while $\boldsymbol{h}_1^t$ of Mode 2 only contains the nearest right blank frames. 
Mode 3 puts right blank frames into the crucial group and leaves trivial group empty. 
Instead, Mode 4 puts left blank frames into the crucial group. 
While Mode 5 inserts both right and left blank frames into the crucial group. 
Among Mode 1 to Mode 5, Mode 2 is verified to be computation efficient with the best recognition performance. 
\begin{table*}[thbp!]
  \caption{ \label{tab:aishell} Comparison of CER(\%), Inv RTF on Aishell-1 datasets without LM. ``Greedy": ``greedy search "; ``Rescoring": ``attention rescoring method". ``Inv-RFT-Greedy(Rescoring)" means  inverse RTF for greedy search(attention rescoring) method. 
 Skipformer\(^\dagger\) uses the checkpoint trained with split Mode 1 as initial checkpoint. }
  \centering
  \begin{tabular}{cccccccc}
    \toprule
     &     &  &  & \multicolumn{2}{c}{Inv-RTF-Greedy\(\uparrow\)}  & \multicolumn{2}{c}{Inv-RTF-Rescoring\(\uparrow\)}  \\
     \cline{5-8}
    Model & Params(M) & Greedy\(\downarrow\)  & Rescoring\(\downarrow\)  & CPU & GPU & CPU & GPU  \\
    \midrule
    Conformer  & 55.2 & 5.12 & 4.64   & 25.8 & 49.6 & 17.1 & 35.4\\
    Efficient   & 48.5 & 4.94  & 4.56  & 24.7 & 57.2 & 19.8 & 53.0\\
    
    \midrule
    Skipformer & 55.2 & 4.48  & 4.27     & \textbf{33.0} & \textbf{81.7} & \textbf{22.8} & \textbf{69.8} \\
    Skipformer\(^\dagger\) & 55.2 & \textbf{4.35}  & \textbf{4.23} & - & - & -  & - \\
    \bottomrule
  \end{tabular}
\end{table*}

\subsection{Intermediate Loss}
We introduce intermediate loss after encoder $E_1$ by using the shared decoder. On the one hand, it can improve the model's performance. 
On the other hand, the output probability of the intermediate CTC will be used as criterion for spliting frames into three groups.  
The frame is labeled as a blank symbol when the posterior probability of the blank symbol for such frame surpasses a pre-defined threshold \(\beta\). 
The total training loss function of our model is 
\begin{equation}
\label{eq_loss}
    \mathcal{L} =\alpha ( \lambda_1 \mathcal{L}_{CTC}^{inter} + \lambda_2 \mathcal{L}_{CTC}^{final}) + (1-\alpha) ( \lambda_1 \mathcal{L}_{AED}^{inter} + \lambda_2 \mathcal{L}_{AED}^{final} )
\end{equation}
where \(\lambda_1\), \(\lambda_2\) and $\alpha$ are hyperparameters. 

\section{Experiments} \label{sect_experiments}
\subsection{Experimental Settings}
We conduct our experiments on  Librispeech~\cite{panayotov2015librispeech} (English) and Aishell-1~\cite{aishell_2017} (Chinese) corpora using U2 model implemented by the  Wenet toolkit~\cite{yao2021wenet}.
For all experiments, the input speech feature is 80-dimentional filterbank (FBank) feature computed on 25ms window with 10ms shift. 
SpecAugment~\cite{specaug} is used for data augmentation. 
All models contain 12 encoder conformer blocks and 6 decoder transformer blocks. 
The decoders for all experiments have the same parameter settings. 
The dimension of encoder and decoder attention layer is 256 with 4 heads, and the dimension of the feed-forward layer is 2048. 
For Aishell-1, we use character modeling units with vocabulary size of 4231. 
And we set $M = 5$ and $N=7$ for encoder $E_1$ and $E_2$.
The Encoder \(E_1\) and \(E_2\) have convolutional kernel size of 15 and 5 respectively. 
For Librispeech, we use SentencePiece for building a 5000 sub-word byte-pair encoding tokenizer. 
The convolutional kernel size is 31 for Encoder \(E_1\) and 9 for Encoder \(E_2\). 
And the architechture has \(M=6\) blocks of \(E_1\) and \(N=6\) blocks of \(E_2\). 
The Adam optimizer [34] with 25000 warmup steps are used for training, where the peak learning rate is 0.001.
We train the models for 300 epochs on Aishell-1 with 4 A100 GPUs and 120 epochs on Librispeech with 8 A100 GPUs. 
For loss weights in equation~\ref{eq_loss}, \(\lambda_1 = 0.5 \), \(\lambda_2 = 0.5 \) and $\alpha=0.3$. 
The blank probability threshold \(\beta\) is set to 0.99 by default. 

We report character error rate (CER) or word error rate (WER) of CTC greedy search and transformer decoder rescoring results for comparison. 
We also report Inverse Real Time Factor (Inv RTF) for measuring computational cost on CPU and GPU devices. 
Inv RTF is measured on the Aishell-1 test set by decoding on one core Intel(R) Xeon(R) Platinum 8350C 2.6GHz CPU with 30G memory or a single A100 GPU. 
On CPU devices, batch size is set to 1.
On GPU devices, batch size is set to 8 for CTC greedy search method and 1 for transformer decoder rescoring. 

\begin{table}[t!]
  \caption{\label{tab:Libri} Comparison of WER(\%) on Librispeech dataset. ``clean(test)" means test-clean(other) testset.}
  \centering
  \begin{tabular}{ccccccccc}
    \toprule
     & \multicolumn{2}{c}{Greedy\(\downarrow\)} & \multicolumn{2}{c}{Rescoring\(\downarrow\)} \\
          \cline{2-5} 
    Model & clean & other & clean & other \\
    \midrule
    Conformer & 3.51 & 9.57 & 3.18 & 8.72 \\
    SqueezeFormer & 3.47 & 8.85  & 3.10 & 8.03 \\

    \midrule
    Skipformer & 3.39 & 8.70 & 3.16  & 8.19 \\
    Skipformer\(^\dagger\) & \textbf{3.20} & \textbf{8.49} & \textbf{3.07}  & \textbf{7.99} \\
    \bottomrule
  \end{tabular}
\end{table}

\subsection{Results on Aishell-1}
Table~\ref{tab:aishell} compares CER on Aishell-1 with recent state-of-the-art models including Conformer\cite{conformer} and  Efficient Conformer\cite{Efficientconformer}. 
``Efficient'' implements downsampling and group attention at different layers by trading off accuracy performance and computational cost. 
All the baseline results can be obtained from \cite{yao2021wenet}.
For both greedy search and transformer decoder rescoring, our proposed Skipformer obtains much better results than vanilla Conformer and Efficient Conformer. 
Skipformer\(^\dagger\) in Table~\ref{tab:aishell} use the pretrain model in Mode 1 as the intial checkpoint to finetune. 
Finally, our proposed Skipformer can obtain 4.23\% CER on Aishell-1 test set and get 8\% relative CER reduction. 
The dowmsampling factor is 8 for both Efficient Conformer and Squeezeformer. 
In contrast, we compute it on test set of Aishell-1 in Mode 2 and its ratio of sequence length of \(\boldsymbol{x}\) to length of \(\boldsymbol{h}_2^c\) reaches approximately to 31 on average. 
Therefore, our model can obtain high Inv RTF during inference. 
As shown in Table~\ref{tab:aishell}, the last four columns show results of Inv RTF on different devices with different methods. 
Inv RTF is tested on the test set of Aishell-1. 
The improvement shows that ignoring redundant information can not only accelerate computation, but also can improve the recognition accuracy. 

\subsection{Results on Librispeech}
Table~\ref{tab:Libri} shows our results on Librispeech dataset with results of SqueezeFormer from ~\cite{squeezeformer} also be reported as an another baseline model. 
The model has comparable numbers of parameters to the baseline model. 
When greedy search method is used, our model outperforms SqueezeFormer significantly with comparable inference speed. As for transformer decoder rescoring method, our model is 47\%/56\% faster than SqueezeFormer on CPU/GPU devices with 
same recognition performance. 
This is due to downsampling factor reaches to 22, which accelerate decoding computation.   

\subsection{Ablation Study on Aishell-1}
Next, we will give various results with different configurations testing on Aishell-1.   

\noindent \textbf{Where to start the frame skipping}. 
We first study effects of number of $E_1$ and $E_2$ layers  on ASR recognition accuracy and the computation complexity. Table~\ref{tab:layer} shows results of models with different $M$ and $N$. It is obvious that the more layers in encoder $E_2$, the less computation the model needs. Therefore, model with $M=2$ and $N=10$ can obtain the higher Inv RTF, especially for the CTC greedy search, because it only relates to the acoustic encoder. 
As \(M=3\) and \(N=9\), the result of CER already surpasses the result of baseline models.  
When $M=5$ and $N=7$, we can get the lowest CER and pretty better Inv RTF.

\begin{table}[ht!]
    \caption{\label{tab:layer} Ablation study of layer numbers of $E_1$ and $E_2$ on Aishell-1 dataset. Mode 2 is adopted for all experiments. $M/N$ represents layer number of encoder $E1$ and $E_2$ separately.}
    \centering
  \begin{tabular}{ccccccc}
    \toprule
    & \multicolumn{3}{c}{Greedy} & \multicolumn{3}{c}{Rescoring}  \\
    \cline{2-7}
    $M/N$ & CER\(\downarrow\) & CPU & GPU & CER\(\downarrow\) & CPU & GPU \\
    
    \midrule
    2/10 & 4.93 & \textbf{36.2} & \textbf{89.2} & 4.68 & \textbf{23.7} & 69.6 \\
    3/9 & 4.65 & 33.2 & 86.6 & 4.41 & 20.5 & \textbf{71.2} \\
    % L-4 & 4.64 & 34.9 & 88.9 & 4.42 & 23.0 & \textbf{72.7} \\
    5/7 & \textbf{4.48} & 33.0 & 81.7 & \textbf{4.27} & 22.8 & 69.8 \\
    % L-6 & 4.81 & 30.5 & 80.1 & 4.56 & 20.6 & 72.1 \\
    % L-7 & 4.53 & 28.9 & 65.6 & 4.32 & 19.5 & 63.3 \\
    8/4 & 4.72 & 25.7 & 56.0 & 4.48 & 17.7 & 67.1 \\
    % L-9 & 4.73 & 28.8 & 87.7 & 4.50 & 20.4 & 74.5 \\
    \bottomrule
  \end{tabular}
\end{table}

\begin{table}[ht!]
    \caption{\label{tab:aishell_mode} Ablation study of split mode on Aishell-1 dataset. }
    \centering
  \begin{tabular}{ccccccc}
    \toprule
    & \multicolumn{3}{c}{Greedy} & \multicolumn{3}{c}{Rescoring}  \\
    \cline{2-7}
    Model & CER\(\downarrow\) & CPU & GPU & CER\(\downarrow\) & CPU & GPU \\
    \midrule
    Mode 1 & 4.49 & 29.8 & 72.1 & 4.31 & 16.7 & 33.1  \\
    Mode 2 & \textbf{4.48} & \textbf{33.0} & \textbf{81.7} & \textbf{4.27} & \textbf{22.8} & \textbf{69.8}  \\
    Mode 3 & 4.53 & 29.7 & 69.8 & 4.30 & 21.5 & 61.2  \\
    Mode 4 & 4.56 & 32.0 & 76.7 & 4.41 & 20.6 & 68.8  \\
    Mode 5 & 4.52 & 28.1 & 58.5 & 4.33 & 20.3 & 63.0  \\
    \bottomrule
  \end{tabular}
\end{table}

\noindent \textbf{Split mode}. 
Next, we will investigate impact of the data split mode on accuracy and inference speed. 
Table~\ref{tab:aishell_mode} shows  results from Mode 1 to Mode 5. As we can see, data split strategy has less impact on accuracy but efficiency. 
However,  the inference speed slow down with more blank frames attended or skipped. 
Attention rescoring method takes great effects on numbers of dropped blank frames. 
Mode 2 includes no blank frames to attend and keeps the least number of blank frames for decoding which takes into account both accuracy and efficiency.

% \subsubsection{Convolutional kernel size}
% \noindent \textbf{Convolutional kernel size} Finally, we try to use smaller  convolutional kernel size at the second stage of encoding. 
% After applying blank skipping method, we find that smaller convolutional kernel size has no effect on CER performance, as shown in Table~\ref{tab:Conv}. 
% \begin{table}[ht!]
%    \caption{\label{tab:Conv} Ablation study of convolutional kernel size at the second stage on Aishell-1 dataset.}
%    \centering
%  \begin{tabular}{ccc}
%    \toprule
%    kernel size & Greedy & Rescoring  \\
%    \midrule
%    15 & 4.54 & 4.39  \\
%    7 & 4.53 & 4.30  \\
%    5 & \textbf{4.48} & \textbf{4.27} \\
%    3 & 4.56 & 4.41  \\
%    \bottomrule
%  \end{tabular}
% \end{table}

\section{Conclusion}\label{sect_conclusion}
In this paper, we propose a ``Skip-and-Recover" method to flexibly downsample acoustic feature at the middle of enocder layers inhomegeneously. 
The blank frames are skipped and ignored for speeding up inference. 
Meanwhile, crucial frames are attended to rest encoder blocks without disturbance of blank features, which promotes the recognition accuracy. 
We also implements several split modes to analysis the effect of blank frames in the encoding and decoding process. 
By splitting frames with different usages, our model shows promising results with lower computational budget. 

\bibliographystyle{IEEEbib}
\bibliography{skipformer}

\begin{thebibliography}{10}

\bibitem{graves2006connectionist}
Alex Graves, Santiago Fern{\'a}ndez, Faustino Gomez, and J{\"u}rgen Schmidhuber,
\newblock ``Connectionist temporal classification: labelling unsegmented sequence data with recurrent neural networks,''
\newblock in {\em Proceedings of the 23rd international conference on Machine learning}, 2006, pp. 369--376.

\bibitem{AED}
Dzmitry Bahdanau, Jan Chorowski, Dmitriy Serdyuk, Philémon Brakel, and Yoshua Bengio,
\newblock ``End-to-end attention-based large vocabulary speech recognition,''
\newblock in {\em 2016 IEEE International Conference on Acoustics, Speech and Signal Processing (ICASSP)}, 2016, pp. 4945--4949.

\bibitem{LAS}
William Chan, Navdeep Jaitly, Quoc Le, and Oriol Vinyals,
\newblock ``Listen, attend and spell: A neural network for large vocabulary conversational speech recognition,''
\newblock in {\em 2016 IEEE International Conference on Acoustics, Speech and Signal Processing (ICASSP)}, 2016, pp. 4960--4964.

\bibitem{graves2012sequence}
Alex Graves,
\newblock ``Sequence transduction with recurrent neural networks,''
\newblock {\em arXiv preprint arXiv:1211.3711}, 2012.

\bibitem{conformer}
Anmol Gulati, James Qin, Chung-Cheng Chiu, Niki Parmar, Yu~Zhang, Jiahui Yu, Wei Han, Shibo Wang, Zhengdong Zhang, Yonghui Wu, and Ruoming Pang,
\newblock ``{Conformer: Convolution-augmented Transformer for Speech Recognition},''
\newblock in {\em Proc. Interspeech 2020}, 2020, pp. 5036--5040.

\bibitem{Efficientconformer}
Maxime Burchi and Valentin Vielzeuf,
\newblock ``Efficient conformer: Progressive downsampling and grouped attention for automatic speech recognition,''
\newblock in {\em 2021 IEEE Automatic Speech Recognition and Understanding Workshop (ASRU)}, 2021, pp. 8--15.

\bibitem{squeezeformer}
Sehoon Kim, Amir Gholami, Albert Shaw, Nicholas Lee, Karttikeya Mangalam, Jitendra Malik, Michael~W Mahoney, and Kurt Keutzer,
\newblock ``Squeezeformer: An efficient transformer for automatic speech recognition,''
\newblock in {\em Advances in Neural Information Processing Systems}, S.~Koyejo, S.~Mohamed, A.~Agarwal, D.~Belgrave, K.~Cho, and A.~Oh, Eds. 2022, vol.~35, pp. 9361--9373, Curran Associates, Inc.

\bibitem{Uconvconformer}
Andrei Andrusenko, Rauf Nasretdinov, and Aleksei Romanenko,
\newblock ``Uconv-conformer: High reduction of input sequence length for end-to-end speech recognition,''
\newblock in {\em ICASSP 2023 - 2023 IEEE International Conference on Acoustics, Speech and Signal Processing (ICASSP)}, 2023, pp. 1--5.

\bibitem{blankregularized}
Yifan Yang, Xiaoyu Yang, Liyong Guo, Zengwei Yao, Wei Kang, Fangjun Kuang, Long Lin, Xie Chen, and Daniel Povey,
\newblock ``Blank-regularized ctc for frame skipping in neural transducer,'' 2023.

\bibitem{wang2023accelerating}
Yongqiang Wang, Zhehuai Chen, Chengjian Zheng, Yu~Zhang, Wei Han, and Parisa Haghani,
\newblock ``Accelerating rnn-t training and inference using ctc guidance,''
\newblock in {\em ICASSP 2023-2023 IEEE International Conference on Acoustics, Speech and Signal Processing (ICASSP)}. IEEE, 2023, pp. 1--5.

\bibitem{FSR}
Zhengkun Tian, Jiangyan Yi, Ye~Bai, Jianhua Tao, Shuai Zhang, and Zhengqi Wen,
\newblock ``{FSR: Accelerating the Inference Process of Transducer-Based Models by Applying Fast-Skip Regularization},''
\newblock in {\em Proc. Interspeech 2021}, 2021, pp. 4034--4038.

\bibitem{factorized}
Xie Chen, Zhong Meng, Sarangarajan Parthasarathy, and Jinyu Li,
\newblock ``Factorized neural transducer for efficient language model adaptation,''
\newblock in {\em ICASSP 2022 - 2022 IEEE International Conference on Acoustics, Speech and Signal Processing (ICASSP)}, 2022, pp. 8132--8136.

\bibitem{u2}
Binbin Zhang, Di~Wu, Zhuoyuan Yao, Xiong Wang, Fan Yu, Chao Yang, Liyong Guo, Yaguang Hu, Lei Xie, and Xin Lei,
\newblock ``Unified streaming and non-streaming two-pass end-to-end model for speech recognition,'' 2021.

\bibitem{panayotov2015librispeech}
Vassil Panayotov, Guoguo Chen, Daniel Povey, and Sanjeev Khudanpur,
\newblock ``Librispeech: an asr corpus based on public domain audio books,''
\newblock in {\em Acoustics, Speech and Signal Processing (ICASSP), 2015 IEEE International Conference on}. IEEE, 2015, pp. 5206--5210.

\bibitem{aishell_2017}
Hui Bu, Jiayu Du, Xingyu Na, Bengu Wu, and Hao Zheng,
\newblock ``Aishell-1: An open-source mandarin speech corpus and a speech recognition baseline,''
\newblock in {\em 2017 20th conference of the oriental chapter of the international coordinating committee on speech databases and speech I/O systems and assessment (O-COCOSDA)}. IEEE, 2017, pp. 1--5.

\bibitem{yao2021wenet}
Zhuoyuan Yao, Di~Wu, Xiong Wang, Binbin Zhang, Fan Yu, Chao Yang, Zhendong Peng, Xiaoyu Chen, Lei Xie, and Xin Lei,
\newblock ``Wenet: Production oriented streaming and non-streaming end-to-end speech recognition toolkit,''
\newblock in {\em Proc. Interspeech}, Brno, Czech Republic, 2021, IEEE.

\bibitem{specaug}
Daniel~S. Park, William Chan, Yu~Zhang, Chung-Cheng Chiu, Barret Zoph, Ekin~D. Cubuk, and Quoc~V. Le,
\newblock ``{SpecAugment: A Simple Data Augmentation Method for Automatic Speech Recognition},''
\newblock in {\em Proc. Interspeech 2019}, 2019, pp. 2613--2617.

\end{thebibliography}

\end{document}